\documentclass[12pt]{article}

\usepackage{amssymb}
\usepackage{amsmath}
\usepackage{graphicx}
\usepackage{hyperref}
\usepackage[square,numbers]{natbib}

\newcommand{\reff}{Eq.~\ref}
\begin{document}

\title{\vspace{-3cm}Achieving interpretable machine learning by functional decomposition of black-box models into explainable predictor effects}

\author{David Köhler$^1$ \and David Rügamer$^{2,3}$ \and Matthias Schmid$^1$}

\date{\small $^1$Institute for Medical Biometry, Informatics and Epidemiology, \\ University of Bonn, Venusberg-Campus 1, D-53127 Bonn, Germany\\
$^2$Department of Statistics, LMU Munich, Ludwigstrasse 33, \\ D-80539 Munich, Germany\\
$^3$Munich Center for Machine Learning, Geschwister-Scholl-Platz 1, \\ D-80539 Munich, Germany
}

\maketitle

\begin{abstract}
\noindent Machine learning (ML) has seen significant growth in both popularity and importance. The high prediction accuracy of ML models is often achieved through complex black-box architectures that are difficult to interpret. This interpretability problem has been hindering the use of ML in fields like medicine, ecology and insurance, where an understanding of the model‘s inner workings is paramount to ensure user acceptance and fairness. The need for interpretable ML models has boosted research in the field of interpretable machine learning (IML). Here we propose a novel approach for the functional decomposition of black-box predictions, which is considered a core concept of IML. The idea of our method is to replace the prediction function by a surrogate model consisting of simpler subfunctions. Similar to additive regression models, these functions provide insights into the direction and strength of the main feature contributions and their interactions. Our method is based on a novel concept termed ``stacked orthogonality‘‘, which ensures that the main effects capture as much functional behavior as possible and do not contain information explained by higher-order interactions. Unlike earlier functional IML approaches, it is neither affected by extrapolation nor by hidden feature interactions. To compute the subfunctions, we propose an algorithm based on neural additive modeling and an efficient post-hoc orthogonalization procedure.
\end{abstract}

\noindent { Keywords: Functional decomposition $|$ Interpretable machine learning $|$ \\ \phantom{Keywords: } Neural additive model $|$ Orthogonality}

\section{Introduction}

{M}achine learning (ML) has increased greatly in both popularity and significance, driven by an increase in methods, computing power and data availability \cite{sarker2021machine}. On July 5, 2024, a search on Web of Science for publications including the term ``machine learning'' yielded more than 350,000 results, corresponding to an average annual increase by more than 20\% since 2006.
ML models are often characterized by their high generalizability, making them particularly successful when used for supervised learning tasks like classification and risk prediction. In recent years, ML models based on deep artificial neural networks (ANNs) have led to groundbreaking results in the development of high-performing prediction models.

The high prediction accuracy of modern ML models is usually achieved by optimizing complex ``black-box'' architectures with thousands of parameters. As a consequence, they often result in predictions that are difficult, if not impossible, to interpret. This interpretability problem has been hindering the use of ML in fields like medicine, ecology and insurance, where an understanding of the model and its inner workings is paramount to ensure user acceptance and fairness. In a recent environmental study, for example, we explored the use of ML to derive predictions of stream biological condition in the Chesapeake Bay watershed of the mid-Atlantic coast of North America~\cite{maloney2022explainable}. Clearly, if these predictions are intended to inform future management policies (projecting, e.g., changes in land use, climate and watershed characteristics), they are required to be interpretable in terms of relevant features as well as the directions and strengths of the feature effects.

\subsection*{Interpretable machine learning}

In recent years, the need for interpretable ML models has boosted research in the field of  interpretable machine learning (IML,~\cite{molnar2020interpretable, murdoch2019}). In this field, {\em interpretability} is commonly defined as ``the degree to which a human can understand the cause of a decision'' \cite{miller2019explanation}. A related concept considered separately in some works is {\em explainability}, which describes ``the internal logic and mechanics that are inside a machine learning system'' \cite{linardatos2020explainable}. Since the methodology presented in this work applies to both concepts, we will not distinguish between the two.

The focus of this paper is on IML for supervised learning tasks, which involve a set of features $X = \{ X_{1}, \ldots , X_d \}$ to derive predictions of a qualitative or quantitative outcome variable $Y$. Denoting the model (i.e., the prediction function) by $F(X) \in\mathbb{R}$, interpretability can generally be achieved in two ways: The first approach is to impose an interpretable structure on $F$ {\em during} the learning process (``model-based interpretability'', \cite{murdoch2019}). A well known example of this approach is the least absolute shrinkage and selection operator \cite{tibshiraniLasso}, which, in its basic form, assumes~$F$ to be linear in the features. Consequently, each feature effect is interpretable in terms of a real-valued coefficient. The second approach, which is particularly applicable to black-box models, aims to achieve interpretability by {\em post}-processing an already learned prediction model (``post hoc interpretability'', \cite{murdoch2019}). Here we will consider \emph{model-agnostic} post-processing methods, which can be applied to a broad range of prediction functions regardless of the ML method applied to the training data~\cite{molnar2020interpretable}. Popular examples of model-agnostic methods include partial dependence plots (PDP) and accumulated local effects (ALE) plots. The underlying principle of these methods is to measure the variability of the prediction function $F$ with respect to changes in subsets of the features~$X$ (an approach that is closely linked to the concept of sensitivity analysis in numerical and nonlinear regression modeling~\cite{da2021basics, lepore:2022}).

While PDP and ALE plots have become established methods in IML, they are not without limitations. For example, PDP have been criticized for ignoring the correlations between the feature of interest and the other features, relying on data points with a very low probability of being observed. This ``extrapolation'' issue may result in misleading effect estimates when the features are correlated~\cite{molnar2020interpretable}. Similarly, PDP may hide possible interaction effects of the features, a problem that can be alleviated by individual conditional expectation (ICE) plots in some cases~\cite{molnar2020interpretable, welchowski}. While ALE plots avoid extrapolation of the data~\cite{apleyZhu}, Gr\"omping~\cite{groemping} observed that these plots do not generally identify the linear shapes of the main effects in a linear prediction model. As a consequence, the feature effects depicted by ALE plots may show systematic deviations from the respective effects in the model formula (for which an explanation is sought). The method proposed in this paper is not affected by these issues: It avoids hiding feature interactions by explicitly including these terms in the estimation procedure and is solely based on the multivariate feature distribution to avoid extrapolation. Furthermore, it does not alter the shapes of the main effects in a linear model.

\subsection*{Functional decomposition}

The basic idea of our method is to achieve interpretability by decomposing the prediction function $F$ (depending on all features $X$) into a set of simpler (``more interpretable'') functions depending on subsets of the features only. More specifically, let $\Upsilon = \{1, \ldots ,d\}$ the set of feature indices and $\mathcal{P}(\Upsilon)$ the power set (i.e., the set of all subsets) of $\Upsilon$. Then $F$ can be decomposed into a sum of~functions
\begin{eqnarray}
    \label{eq:GeneralDecomp0}
    \hspace{-.2cm}F(X) \hspace{-.1cm} & \hspace{-.1cm} = & \hspace{-.1cm}\mu \, + \sum_{\theta\in \mathcal{P}(\Upsilon): |\theta| = 1} f_{\theta}(X_{\theta}) \, +
\sum_{\theta\in\mathcal{P}(\Upsilon): |\theta| = 2} f_{\theta}(X_{\theta}) \nonumber\\
&& \hspace{-.1cm} +\,
\ldots \, + \sum_{\theta\in\mathcal{P}(\Upsilon): |\theta| = d} f_{\theta}(X_{\theta}) \, ,
\end{eqnarray}
where $\mu \in\mathbb{R}$ is an intercept term and, for any $\theta \in \mathcal{P}(\Upsilon)\backslash\emptyset$, $X_\theta$~denotes the subset of features with indices $\theta$. For example, if $d=3$ and $\theta = \{1,3\}$, then $X_\theta$ is given by $\{X_1, X_3 \}$. Accordingly, the intercept term can be defined as $\mu = f_\emptyset$. Note that the last sum in~\reff{eq:GeneralDecomp0} consists of only one summand.

In IML, the main focus is usually on the subset of functions~$f_\theta$ with $|\theta| = 1$ (``main effects'', first sum in \reff{eq:GeneralDecomp0}) and $|\theta| = 2$ (``two-way interactions'', second sum in \reff{eq:GeneralDecomp0}). For main effects, $f_\theta$ depends on only one feature $X_j$, $j \in \Upsilon$, allowing for a simple graphical analysis that plots the values of $f_\theta (X_j)$ against the values of $X_j$. Two-way interactions, on the other hand, can be visualized using heatmaps or contour plots. Both main effects and two-way interactions allow for simple graphical interpretations of the respective feature effects, whereas the functions $f_\theta$ with $|\theta| > 2$ (termed ``multivariate feature interactions'') constitute the less interpretable parts of $F$.

This paper presents a novel approach to specify and compute the functions $f_\theta$, given a fixed (possibly black-box) prediction function $F$. The proposed method also allows the measurement of the ``degree of interpretability'' by quantifying the importance of the main and two-way interaction effects in \reff{eq:GeneralDecomp0}. We emphasize that our methodology is designed to {\it decompose} the prediction function $F$ but not to {\it learn} it from a set of data. Accordingly, we assume that $F$ is not subject to sampling variability but has been derived previously by the application of some ML method. Our method is based on regularity conditions that are similar to those described by Hooker \cite{hooker2007generalized}; however, we consider a different type of functional decomposition and also employ a different computational methodology.

\section{Methods}

\subsection*{Conditions on the features and the prediction function}

It is clear from \reff{eq:GeneralDecomp0} that the functions $f_\theta$ are not uniquely defined. For example, let $d=2$, $\mu=0$ and $F(X_1,X_2) = X_1 + X_1\cdot X_2$. Then the sets of functions $\{ f_1(X_1)=X_1, f_2(X_2) = 0$, $f_{12}(X_1,X_2) = X_1\cdot X_2\}$ and $\{f_1(X_1)=0.5\cdot X_1 , f_2(X_2) = 0$, $f_{12}(X_1,X_2) = 0.5\cdot  X_1 + X_1\cdot X_2 \}$ both satisfy \reff{eq:GeneralDecomp0}. As a consequence, further assumptions are needed to derive a unique representation of \reff{eq:GeneralDecomp0}.

Our first set of assumptions is on the features $X= \{X_1, \ldots , X_d\}$. In line with Hooker \cite{hooker2007generalized}, we consider the features as real-valued random variables, assuming that $X_1, \ldots , X_d$ are defined on a joint probability space with probability function~$P_X$. We further assume that each~$X_j$, $j\in\Upsilon$, has bounded support. 
Note that these are rather weak assumptions in practice, allowing~$X$ to include both continuous and categorical features (the latter encoded by sets of dummy variables).

Regarding the functions in \reff{eq:GeneralDecomp0}, we assume that each~$f_\theta$, and also $F$, is square integrable with respect to $P_X$. Again, this is a rather weak assumption, as square integrable functions emerge from many popular ML methods. They include, for instance, the piecewise prediction functions obtained from random forests and tree boosting, and also many ANN predictors after transformation by a sigmoid activation function. Following Hooker \cite{hooker2007generalized}, we define the variance of~$f_\theta$, $\theta\in\mathcal{P}(\Upsilon)\backslash\emptyset$, by $\sigma_\theta^2 = \int f_\theta^2 (X_\theta ) dP_{X}$, the variance of $F$ by $\sigma_F^2 = \int (F (X) - \mu)^2 dP_{X}$, and the covariance of $f_\theta$ and $f_{\theta'}$, $\theta , \theta' \in \mathcal{P}(\Upsilon)\backslash\emptyset$, by $\sigma_{\theta\theta'} = \int f_\theta (X_\theta )f_{\theta'} (X_{\theta'} ) dP_{X}$. Without loss of generality, we assume that each $f_\theta$, $\theta\in \mathcal{P}(\Upsilon)\backslash\emptyset$, is centered around zero, i.e.\@ $\int f_\theta (X_\theta) dP_{X} = 0$ (\cite{hooker2007generalized}, p.\@~714). Finally, we assume that the functions $f_\theta$, $\theta\in \mathcal{P}(\Upsilon)\backslash\emptyset$, are linearly independent. This assumption means that each~$f_\theta$ forms a closed subspace of the Hilbert space of square integrable functions. In practice, it implies that each $f_\theta$ carries unique information about $F$ and that all functions $f_\theta$ are non-zero.

\subsection*{Generalized functional ANOVA}

Next we define a set of requirements to describe the relations between the functions~$f_\theta$. Our main requirement is that the summands in \reff{eq:GeneralDecomp0} are well separated, meaning that higher-order effects (i.e.\@ functions with large $| \theta |$) do not contain any components of lower-order effects with small $| \theta |$ (see below for a mathematical treatment). In particular, we require that predictive information explained by a main effect is not contained in the higher-order effects that include the corresponding feature ({\em purity} criterion, Molnar~\cite{molnar2020interpretable}, Section~8.4). A related requirement is {\em optimality}, meaning that lower-order functions should capture as much functional behavior as possible \cite{hooker2007generalized}.

To implement the above requirements, Hooker \cite{hooker2007generalized}  proposed a decomposition termed {\em generalized functional ANOVA}. With this approach, the functions in \reff{eq:GeneralDecomp0} are required to be {\em hierarchically orthogonal}, satisfying the constraints
\begin{equation}
\label{eq:hookerOrthog} \hspace{-.13cm}
\forall \theta \in \mathcal{P}(\Upsilon)\backslash\emptyset \ \
\forall \theta' \subseteq \theta \hspace{-.04cm} : \,
\sigma_{\theta\theta'} \hspace{-.05cm} = \hspace{-.05cm}\int\hspace{-.05cm} f_\theta (X_\theta )f_{\theta'} (X_{\theta'} ) dP_{X} \hspace{-.05cm}= \hspace{-.05cm}0 \, .
\end{equation}
Hierarchical orthogonality implies that for any given $\theta'$, the effect $f_{\theta'}(X_{\theta'})$ is orthogonal to all higher-order effects~$f_\theta (X_{\theta})$ with $X_{\theta} \supseteq X_{\theta'}$ \cite{hooker2007generalized, stone1994}. It thus provides an implementation of the purity criterion, ensuring that higher-order effects are uncorrelated with lower-order effects. Furthermore, the constraints in \reff{eq:hookerOrthog} provide an implementation of optimality because, according to \reff{eq:hookerOrthog}, all lower-order effects~$f_{{\theta'}}$ are orthogonal projections of the combined effects $f_{\tilde{\theta}}:= f_{{\theta'}} + f_{{\theta}}$ onto the respective lower-order subspaces. It follows from the Hilbert projection theorem \cite{luenberger1969} that the lower-order effects~$f_{{\theta'}}$ capture as much of the variance of $f_{\tilde{\theta}}$ (i.e.\@ as much functional behavior of $f_{\tilde{\theta}}$) as possible.

In his original work on generalized functional ANOVA, Hooker \cite{hooker2007generalized} specified conditions for the uniqueness of the~functions $f_\theta$\footnote{Note that Hooker considered a more general definition of the integral in \reff{eq:hookerOrthog}, allowing for weight functions other than the probability density function of $X$.}.
Based on the same decomposition, Chastaing et~al.~\cite{chastaing2012generalized} studied further assumptions on the feature distribution~$P_X$. The authors also introduced a coefficient to measure the importance of individual feature combinations. For each~$\theta \in \mathcal{P}(\Upsilon)\backslash\emptyset$, this coefficient is defined as $S_\theta = (\sigma_\theta^2 + \sum_{\theta' \neq \theta} \sigma_{\theta\theta '}) / \sigma_F^2$ (\textit{generalized Sobol sensitivity index}, Chastaing et al.\@~\cite{chastaing2012generalized}, p.\@~2427).

\subsection*{Computational challenges}
In recent years, functional decomposition has been acknowledged as a key concept in making ML models explainable \cite{molnar2020interpretable}. In practice, however, the application of functional decomposition methods remains challenging. This is mainly due to the computational and numerical issues associated with the estimation of the feature effects $f_\theta$. In fact, despite the availability of algorithms to achieve hierarchical orthogonality \cite{chastaing2012generalized, hooker2007generalized, lengerich}, state-of-the-art methods still involve systems of equations that are, even for a moderate feature count, ``complex and computationally intensive'' \cite{molnar2020interpretable}. In the following, we will introduce {\em stacked orthogonality}, an alternative approach to implement purity and optimality. Based on the conditions of stacked orthogonality, we will present an algorithm to estimate the functions $f_\theta$ in a computationally efficient manner.

\subsection*{Functional decomposition with stacked orthogonality}

Analogous to generalized functional ANOVA, our method is based on the functional decomposition in \reff{eq:GeneralDecomp0}. However, instead of the hierarchical orthogonality constraints in \reff{eq:hookerOrthog}, we require the functions $f_\theta$ to meet the {\em stacked orthogonality} constraints
\begin{eqnarray}
\label{eq:stackedOrthog}
&& \hspace{-1.05cm}\forall k \in \Upsilon \hspace{-.05cm} : \hspace{-.05cm} \int \hspace{-.05cm} \Big(\sum_{
\overset{\theta \in \mathcal{P}(\Upsilon):}{|\theta| = k}}f_\theta (X_\theta ) \Big) \Big(\sum_{
\overset{\theta' \in \mathcal{P}(\Upsilon):}{|\theta'| < k}} f_{\theta'} (X_{\theta'} ) \Big)
dP_{X} = 0 \, ,
\end{eqnarray}
where $k \in \Upsilon$ denotes the effect level\footnote{In the following, we will use the terms ``order'' and ``level'' interchangeably.}. Unlike hierarchical orthogonality, which requires the effect of each {\em individual} feature combination $\theta$ to be uncorrelated with higher-order effects, the conditions in \reff{eq:stackedOrthog} provide a {\em level-wise} implementation of the purity criterion: For each level $k$, the sum of all level-$k$ effects is required to be uncorrelated with the sum of all lower-level effects (including the intercept with $|\theta'|=0$) -- hence the term ``stacked orthogonality''. In addition to implementing purity, the constraints in \reff{eq:stackedOrthog} also provide a level-wise implementation of optimality. This is because, according to \reff{eq:stackedOrthog}, the sum of the ``lower-order'' effects (with levels $<k$) is an orthogonal projection of the sum of the ``current-order'' effects (with levels~$\le k$) onto the lower-order subspace. It follows from the Hilbert projection theorem that the sum of lower-order effects captures as much of the variance of the sum of the current-order effects (i.e.\@ as much functional behavior at the current level) as possible. Note that assuming linear independence of the functions in \reff{eq:GeneralDecomp0} guarantees the closedness of the lower-order subspaces, and also a unique representation of the effects $f_\theta$, $|\theta | = k$, at each level $k \in \Upsilon$.

A convenient feature of stacked orthogonality is that the variance of $F$ can be decomposed in a level-wise fashion, giving rise to the calculation of level-wise coefficients of explained variation. More specifically, for each $k \in \Upsilon$, we define the {\em fraction of $\sigma_F^2$ explained by the $k$-th level} as
\begin{equation} \label{Ik}
I_k = \frac{\int
\left(\sum_{\theta' \in \mathcal{P}(\Upsilon): |\theta'| = k}f_{\theta'} (X_{\theta'} ) \right)^2
dP_{X}}{\sigma_F^2} \, .
\end{equation}
Consequently, by calculating $I_1$ (fraction of $\sigma_F^2$ explained by the main effects) and $I_2$ (fraction of $\sigma_F^2$ explained by the two-way interaction effects), it is possible to quantify the {\em degree of interpretability} of the prediction model~$F$. We emphasize that the definition in \reff{Ik} is different from the generalized Sobol sensitivity indices in~\cite{chastaing2012generalized}, as the latter refer to contributions of individual feature combinations~$\theta$, $\theta \in \mathcal{P}(\Upsilon )$, whereas $I_k$, $k\in \Upsilon$, measure the level-wise contributions of all features.

\subsection*{Estimation by neural additive models and post-hoc orthogonalization}

As stated above, the application of functional decomposition methods strongly depends on the availability of a user-friendly algorithm to compute the functions $f_\theta$. We propose the following three-step procedure to arrive at the decomposition in \reff{eq:GeneralDecomp0} satisfying the stacked orthogonality constraints in~\reff{eq:stackedOrthog}:

{\em In the first step}, we generate $n$ data points $\mathcal{S} = \{ F_i, X_{i1}, \ldots , X_{id} \}_{i = 1, \ldots , n}$, where $X_{ij}$, $j \in \Upsilon$, and $F_i = F(\{ X_{i1}, \ldots , X_{id} \})$ denote the $j$-th feature value and the value of the prediction function, respectively, of the $i$-th data point. For instance, the data could be sampled from an available set of training data that were used previously for the learning of $F$. In this case, the probability measure~$P_X$ is given by the distribution of the feature values in the training data. Alternatively, one could use a grid of feature values to generate $\mathcal{S}$ (corresponding to uniformly distributed features) or some other reference distribution for which an explanation is sought.

\begin{figure}
    \centering
    \includegraphics[width = 0.9\textwidth]{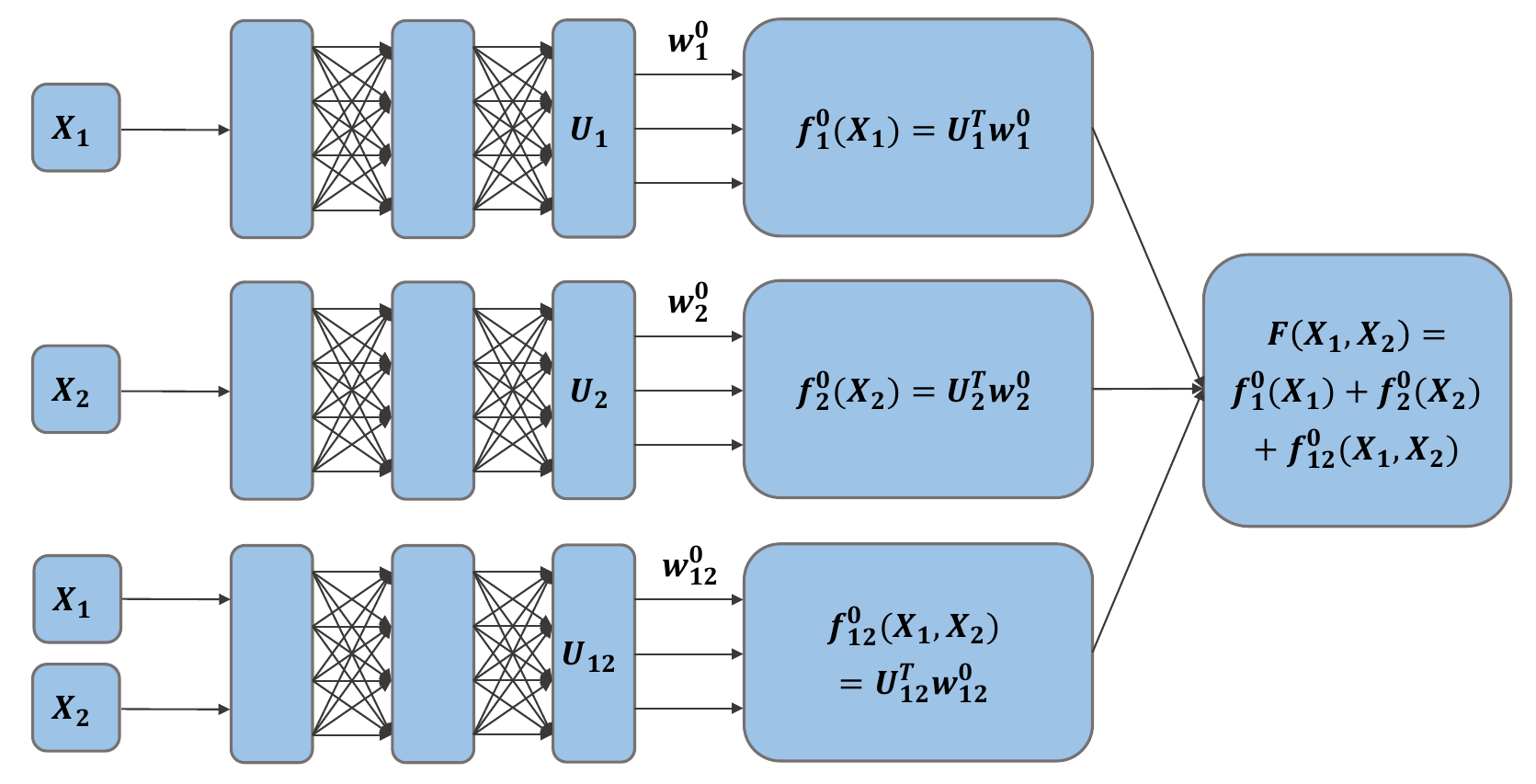}
    \caption{Illustration of the neural additive model in \reff{eq:NAM}. In the example considered here, there are two features $X_1$ and $X_2$. Accordingly, the set of functions $f_\theta^0$, $\theta\in\mathcal{P}(\Upsilon)\backslash\emptyset$, is given by the two main effects $f_1^0(X_1)$, $f_2^0 (X_2)$ and the two-way interaction~$f_{12}^0(X_1, X_2)$. Each function is represented by a fully connected artificial neural network (ANN). The units in the penultimate layers of the ANNs are denoted by $U_1 \in \mathbb{R}^{b_1}$, $U_2  \in \mathbb{R}^{b_2}$ and $U_{12} \in \mathbb{R}^{b_{12}}$, where $b_1$, $b_2$ and $b_{12}$ are the widths of the layers. The outputs of the ANNs are given by the dot products $U_1^\top \mathbf{w}_1^0$, $U_2^\top \mathbf{w}_2^0$ and $U_{12}^\top \mathbf{w}_{12}^0$, where $\mathbf{w}_1^0$, $\mathbf{w}_2^0$ and $\mathbf{w}_{12}^0$ are vectors of weights. The prediction function $F(X_1, X_2)$ is given by the sum of the three dot products (hence the term neural {\em additive} model). The parameters of the ANNs are estimated jointly by backpropagation. Details on model fitting and the specification of the ANN architectures are given in Appendix A.}
    \label{fig:DNN}
\end{figure}

{\em In the second step}, we use the data generated in Step 1 to obtain initial estimates ${f}_{\theta}^0$ of the functions  $f_\theta$, $\theta \in \mathcal{P}(\Upsilon)\backslash\emptyset$. This is done by fitting a {\em neural additive model} (NAM, \cite{agarwalNAM}) of the form
\begin{eqnarray}
    \label{eq:NAM}
    \hspace*{-0.4cm} F_i &=& \sum_{\theta\in \mathcal{P}(\Upsilon): |\theta| = 1} {f}_{\theta}^0 (X_{i\theta})+
\sum_{\theta\in\mathcal{P}(\Upsilon): |\theta| = 2} {f}_{\theta}^0  (X_{i\theta}) \nonumber \\ && +
\, \ldots \, + \sum_{\theta\in\mathcal{P}(\Upsilon): |\theta| = d} {f}_{\theta}^0  (X_{i\theta}) \, , \ \ \ i=1,\ldots ,n \, ,
\end{eqnarray}
where $X_{i\theta}$ are the values of $X_\theta$ corresponding to the $i$-th data point. Model fitting is performed using a backpropagation procedure, with each function~${f}_\theta^0$ represented by an ANN depending on the respective feature subset $X_\theta$ (see Figure~\ref{fig:DNN} for an illustration). As demonstrated by Agarwal et al.\@ \cite{agarwalNAM}, NAMs allow for modeling a wide range of functional shapes, exploiting the property of ANNs to approximate general classes of functions arbitrarily well \cite{cai2023, cybenko1989, hornik1991, hornik1989, ismailov2022, ismailov2023, kidger, park2021minimum, shen2022}. Compared to Agarwal et al.\@ \cite{agarwalNAM}, our only additional requirement (needed for Step 3 below) is that all ANNs in \reff{eq:NAM} are linear in their output layers. More specifically, we require each vector\linebreak $\mathbf{{f}}_\theta^0  = ({f}_\theta^0  (X_{1\theta}), \ldots , {f}_\theta^0  (X_{n\theta}))^\top \in \mathbb{R}^n$, $\theta \in \mathcal{P}(\Upsilon)\backslash\emptyset$, to be of the form
\begin{equation} \label{eq:lastLayerLinear}
\mathbf{{f}}_\theta^0 = \mathbf{U}_\theta \mathbf{w}_\theta^0 \, ,
\end{equation}
where $\mathbf{U}_\theta \in \mathbb{R}^{n \times b_\theta}$ and $b_\theta \in \mathbb{N}$ are the outputs and the number of units, respectively, of the penultimate layer, and $\mathbf{w}_\theta^0 \in \mathbb{R}^{b_\theta}$ is a vector of weights. Note that \reff{eq:NAM} does not contain an intercept term. Accordingly, the initial estimate of $\mu$ is given by $\mu^0 = f_\emptyset^0 = 0$, and we define $b_\emptyset = 1$, $\mathbf{U}_\emptyset = (1,\ldots ,1)^\top \in\mathbb{R}^{n\times 1}$, and $\mathbf{w}_\emptyset^0 = 0$. Updates of the initial intercept vector $\mathbf{f}_\emptyset^0 = \mathbf{U}_\emptyset \mathbf{w}_\emptyset^0 \in\mathbb{R}^n$ will be computed during the post-hoc orthogonalization procedure described below. 
For details on the specification of the ANNs in \reff{eq:NAM}, we refer to Appendix A.

We emphasize that we do not use NAMs for supervised learning, i.e.\@ to derive the relationship between an outcome variable $Y$ and a set of features $X_1, \ldots , X_d$. Instead, we consider the predicted values $F_i$ as the outcome of the NAM. Given $X_i$, these values are deterministic, and hence the right-hand side of \reff{eq:NAM} does not include a residual error term. Put differently, the right-hand side of \reff{eq:NAM} defines a ``surrogate model'' for the prediction model~$F$. Importantly, since we want to arrive at an exact decomposition of the form \reff{eq:GeneralDecomp0}, we do {\em not} aim to avoid overfitting the data. Instead, we run the backpropagation procedure until it achieves an (almost) perfect correlation between the left-hand side and the right-hand side of \reff{eq:NAM}. This is possible due to the approximation properties of ANNs (see the section on experiments with synthetic data). We further note that model fitting can be done very conveniently using established ANN implementations in Python and R.

{\em In the third step}, we apply a {\em post-hoc orthogonalization procedure} to the initial estimates $f_\theta^0$. This is necessary to ensure that the final estimates satisfy the stacked orthogonality conditions in \reff{eq:stackedOrthog}\footnote{Note that the NAM architecture in Figure~\ref{fig:DNN} does not guarantee stacked orthogonality of the functions~$f_\theta^0$.}. The post-hoc orthogonalization procedure considered here is an extension of the method by R\"ugamer~\cite{ruegamer2023}; it proceeds in an iterative manner, starting at the highest interaction level and descending down to the main effects. We describe the first two iterations of the procedure in a non-technical way. A formal definition of the algorithm is given in Appendix B.

In the first iteration of the post-hoc orthogonalization procedure, the idea is to achieve orthogonality between the $d$-way interaction effect and the sum of all lower-order effects ($|\theta | < d$). To this end, the vector of $d$-way interactions (given by $\mathbf{{f}}_{\Upsilon}^0$) is projected onto the column space spanned by the ``lower-order'' matrices $\mathbf{U}_\theta$, $|\theta| < d$ (including $\mathbf{U}_\emptyset$, which is a vector of ones). Next, $\mathbf{{f}}_{\Upsilon}^0$~is replaced by the vector orthogonal to this space, giving the new vector of $d$-way interactions~$\mathbf{{f}}_{\Upsilon}^1$. Note that $\mathbf{{f}}_{\Upsilon}^1$ has zero mean, as the lower-order column space contains a column of ones, and as $\mathbf{{f}}_{\Upsilon}^1$ is orthogonal to this space. The lower-order functions are updated by adding the projected values of~$\mathbf{{f}}_{\Upsilon}^0$ to the initial lower-order functions, giving new functions~$\mathbf{{f}}_{\theta}^1$, $|\theta| < d$ (including a new intercept~$\mathbf{f}_\emptyset^1$).

In the second iteration, the idea is to achieve orthogonality between the sum of the effects of order $d-1$ and the sum of all effects with $|\theta| < d-1$. Analogous to the first iteration, the effects of order $d-1$ are summed up and projected onto the column space spanned by the matrices $\mathbf{U}_\theta$, $|\theta| < d-1$ (again including $\mathbf{U}_\emptyset$). Next, each $\mathbf{{f}}_{\theta}^1$ with $|\theta| = d-1$ is replaced by its respective vector orthogonal to this column space, giving new estimates~$\mathbf{{f}}_{\theta}^2$ of the effects of order $d-1$. The functions with $\theta < d-1$ are updated in the same way as in the first iteration, resulting in new estimates $\mathbf{{f}}_{\theta}^2$, 
$|\theta| < d-1$, whereas the ``higher-order'' vector $\mathbf{{f}}_{\Upsilon}^1$ is left unchanged ($\mathbf{{f}}_{\Upsilon}^1 \equiv \mathbf{{f}}_{\Upsilon}^2$).

It is clear that iterating the above procedure (i.e., establishing orthogonality between the sums of the current-order and the lower-order effects while leaving higher-order effects unchanged) ensures stacked orthogonality of the final estimates $\mathbf{{f}}_{\theta}^{d-1}$. As a result, one obtains the desired decomposition of the prediction function~$F$. We emphasize that post-hoc orthogonalization does not require re-fitting the NAM in \reff{eq:NAM} but can be performed rather efficiently by multiplying a set of matrices and vectors. In case of a high(er)-dimensional feature set, the number of summands in \reff{eq:NAM} can easily be reduced to a subset of ``relevant'' effects, see Remark~2 below. Our method is implemented in Python and~R.

{Remark 1:} NAM fitting is based on ANN layers with prespecified numbers of hidden units. We note that these numbers may not always be sufficient to closely approximate the true underlying functions, especially when the latter are highly non-linear. To address this issue and to ``further improve accuracy and reduce the high-variance that can result from encouraging the model to learn highly non-linear functions'', Agarwal et al.\@ \cite{agarwalNAM} proposed to compute the final function estimates by an average of multiple NAM fits (``ensemble approach''). In line with this strategy, we stabilize our function estimates by fitting an ensemble of NAMs with different weight initializations and by applying the post-hoc orthogonalization procedure to each member of the ensemble. Afterwards the orthogonalized estimates are averaged, giving vectors of the form $\bar{\mathbf{f}}_\theta^{d-1} = \sum_{r=1}^R \mathbf{f}_\theta^{d-1, r} / R$, where $R$ is the size of the ensemble and $\mathbf{f}_\theta^{d-1, r}$ refers to the post-hoc-orthogonalized estimate of the $r$-th ensemble member. Note that this procedure does not substantially increase the run time of the algorithm, as NAM fitting with different weight initializations can be parallelized. We further note that the averaged estimates are no longer guaranteed to satisfy the stacked orthogonality constraints in \reff{eq:stackedOrthog}. To overcome this problem, we add a final post-hoc orthogonalization step to our algorithm, replacing the outputs~$\mathbf{U}_\theta$ by the averaged vectors~$\bar{\mathbf{f}}_\theta^{d-1}$ and applying the above procedure to the averaged estimates. 

{Remark 2:} In settings with a large number of features, the number of interaction terms in \reff{eq:NAM} is very high ($\sum_{l=2}^{d}\binom{d}{l}$, growing exponentially in~$d$). In these cases, one is often interested in the interpretation of only a small subset of effects. The stacked orthogonality approach can easily be adapted to these settings; all one has to do is to redefine the NAM in Step 2. To this end, let $\Theta \subset \mathcal{P}(\Upsilon )\backslash\emptyset$ represent the effects of interest, and let $\mathcal{P}(\Upsilon ) \backslash (\Theta \cup \emptyset) $ be the corresponding set of ``non-interesting'' effects. Then $\mathcal{P}(\Upsilon ) \backslash (\Theta \cup \emptyset)$ can be removed from the lower-order sums in~\reff{eq:NAM} and absorbed into the last summand $f_{\Upsilon}^0$. Post-hoc orthogonalization can be applied to the resulting NAM fit as before.

{Remark 3:} The coefficient $I_k$ can be computed by replacing the variance terms in \reff{Ik} with their respective sample variances obtained from the post-hoc-orthogonalized ensemble average.

\section{Experiments with synthetic data}

We investigated whether our method is able to extract the subfunctions $f_\theta$ from a synthetic additive prediction function. To this end, we constructed predictions defined by
\begin{eqnarray}
\label{eq:simModel}
\hspace{-.4cm} F_i \hspace{-.2cm} &= \hspace{-.2cm}& f_1 (X_{i1}) + f_3 (X_{i3}) + f_3 (X_{i3}) \nonumber\\
&& + \ f_{12} (X_{i1},X_{i2}) + f_{13} (X_{i1},X_{i3}) + f_{23} (X_{i2},X_{i3})\nonumber\\
&& + \ f_{1,\ldots,10}(X_{i1},\ldots ,X_{i10}) \, ,
\end{eqnarray}
where $X_1,\ldots , X_{10}$ followed a multivariate uniform distribution on $[-3,3]^{10}$. In our experiments, we considered three scenarios with different sets of functional forms for the main and two-way interaction effects (for details, see Appendix C). In order to define the true  decomposition that our method should recover, we orthogonalized these functions in a large data set of size $n=100,000$. Using the obtained orthogonal functions, we generated 10 independent samples $\{ F_i, X_{i1}, \ldots , X_{i10} \}_{i = 1, \ldots , n}$ of size $n\in \{2000,5000\}$ to which we applied our method. The feature values were generated by sampling data points $\{ Z_{i1}, \ldots , Z_{i10} \}_{i = 1, \ldots , n}$ from a multivariate normal distribution with zero mean, unit variance and equicorrelation~$0.5$, and by applying the univariate standard normal cumulative distribution function $\Phi (\cdot )$ to give $X_{ij} = 6 \cdot (\Phi (Z_{ij}) -0.5)$, $j=1,\ldots , 10$. We used the ANN architecture described in Appendix A, setting the number of ensemble members to 10.

\begin{figure}[h!]
    \centering
    \includegraphics[width = 0.7\textwidth]{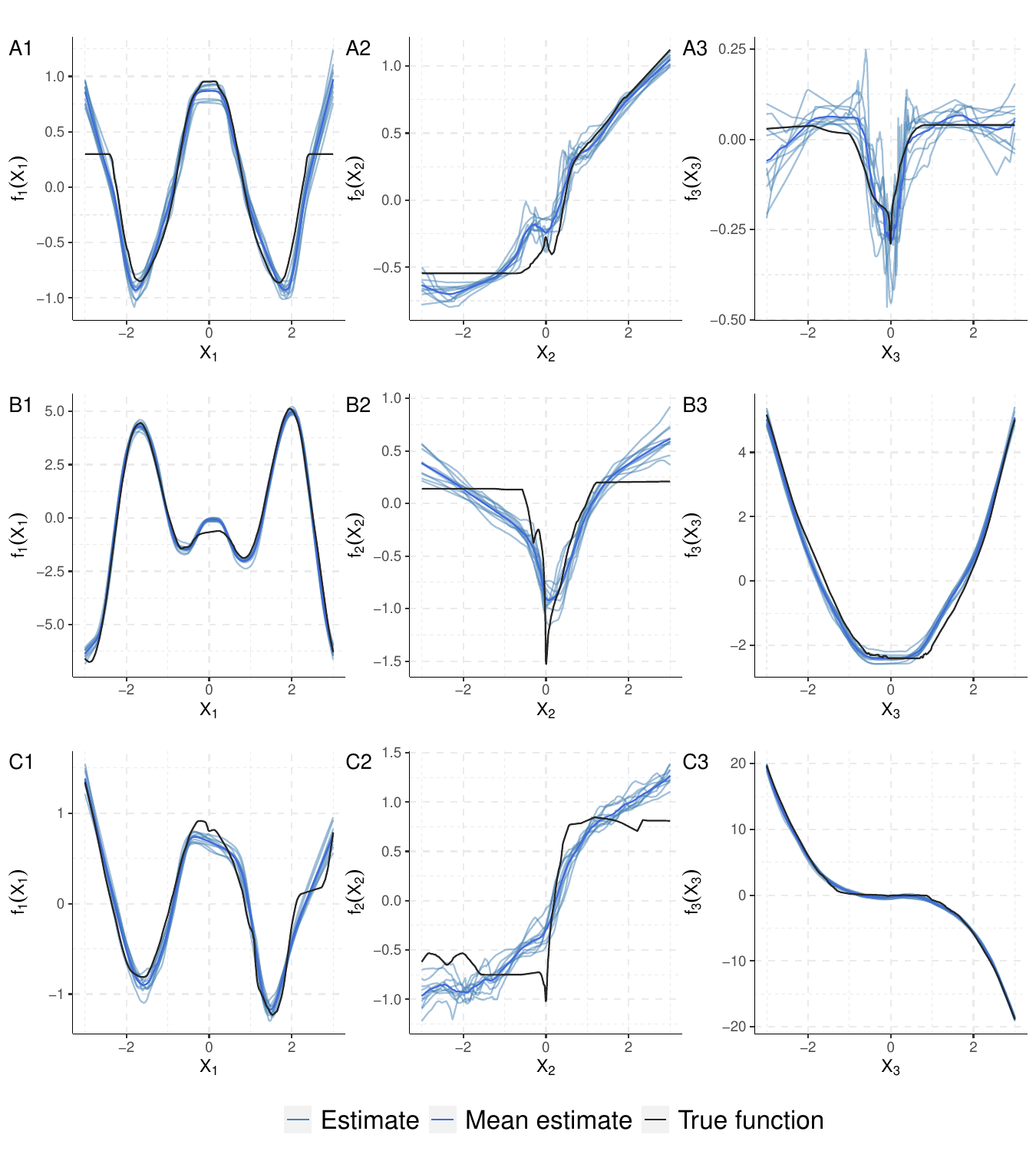}
    \caption{Experiments with synthetic data. The blue lines visualize the main effects $f_1(X_1), f_2(X_2), f_3(X_3)$, as obtained by applying the proposed three-step algorithm to samples of size $n=2000$ each. The black lines correspond to the true post-hoc-orthogonalized main effects defined in Appendix~C \linebreak (A1-A3: scenario 1, B1-B3: scenario 2, C1-C3: scenario 3).}
     \label{sim:fig2000main}
\end{figure}

Figure \ref{sim:fig2000main} presents the estimated main effects obtained in the three scenarios with $n=2000$. Despite some variation, which is likely due to differences in the empirical distribution functions of the features, and some tendency to oversmooth the effects in highly nonlinear regions (which could be addressed by increasing the complexity of the NAM architecture), our method performed well in approximating the true main effects. Similar results were obtained for the two-way interaction terms and in the scenarios with $n=5000$ (Appendix~D).

The average values of the summary measures $I_1$ and $I_2$ were $0.511,0.924$, $0.983$ and $0.412,0.073,0.017$, respectively, in the scenarios with $n=2000$, and $0.502,0.921,0.980$ and $0.475,0.077,0.019$, respectively, in the scenarios with $n=5000$.\\ \\

\clearpage

\section{Discussion}

In recent years, techniques to improve the interpretability of black-box models have become a key component of machine learning methodology. As part of this methodology, functional decomposition is considered a ``core concept of machine learning interpretability'' \cite{molnar2020interpretable}.

In this paper we proposed a novel concept for the decomposition of black-box prediction functions into explainable feature effects. In line with earlier approaches by Hooker \cite{hooker2007generalized}, the idea of our method is to replace the original prediction function by a surrogate model consisting of simpler ``more interpretable'' subfunctions. The latter allow for a graphical representation of the main feature contributions and their interactions, providing insights into the direction and strength of the effects. 

Our concept of  stacked orthogonality is designed to achieve purity of the subfunctions; it implies that predictive information explained by the main effects is not contained in the higher-order effects. At the same time, stacked orthogonality implies that lower-order functions (offering a high degree of interpretability) capture as much functional behavior as possible. Another contribution of this work is development of a user-friendly algorithm to estimate the subfunctions from data. It is based on the fitting of a neural additive model (NAM), which allows the approximation of feature effects using ANN architectures, and an efficient post-hoc orthogonalization method to achieve stacked orthogonality. The proposed algorithm was able to approximate the true underlying subfunctions in our numerical experiments.

Despite the aforementioned advantages, our method is not without limitations. First, NAM fitting (and thus estimation of the subfunctions) is limited to rather low-dimensional feature sets. In particular, it cannot be done with arbitrarily large numbers of interaction terms. It should be emphasized, however, that our method allows users to specify subsets of ``effects of interest'' and to shift all ``uninteresting'' effects to the highest-order interaction level. This strategy preserves the practicability of the proposed method even when the overall number of higher-order interactions is prohibitively large. A second limitation is that our concept of stacked orthogonality is not primarily designed for quantifying the overall contributions of single features. Instead, our summary measures $I_k$ quantify the contributions of the effect {\em levels} (e.g., all main effects or all interaction effects considered together), or more generally, the contributions of the aforementioned ``effects of interest'' to the overall black-box prediction. On the other hand, our method does not preclude users from calculating generalized Sobol sensitivity indices (as defined by Chastaing et al.\@ \cite{chastaing2012generalized}) to summarize the overall contributions of single features.

Finally, we emphasize that our method is designed to explain the {\em inner} workings of a black-box model. It can {\em not} be used to evaluate the features' ability to predict the outcome variable $Y$. This is a general aspect of post-hoc functional decomposition \cite{chastaing2012generalized, hooker2007generalized, molnar2020interpretable} and can be deduced from the basic equation in \reff{eq:GeneralDecomp0}. In fact, since the left-hand side of \reff{eq:GeneralDecomp0} is entirely dependent on the prediction function~$F$ but not on~$Y$, the decomposition in \reff{eq:GeneralDecomp0}, and thus also the NAM in~\reff{eq:NAM}, do not incorporate any information on how well $Y$ can be predicted by $F$ and its subfunctions $f_\theta$. Put differently, the subfunctions obtained from our method will only have a useful {\em interpretation} if the underlying black-box model is useful in {\em predicting} the outcome of interest. \\ \vspace{.2cm}

\noindent {\bf Acknowledgements}\\ \vspace{-.4cm}

\small
\noindent We thank Thomas Prince for proofreading the manuscript. \vspace{.3cm}

\vspace{.3cm}

\bibliography{references}

\begin{thebibliography}{10}

\bibitem{agarwalNAM}
R.~Agarwal, L.~Melnick, N.~Frosst, X.~Zhang, B.~Lengerich, R.~Caruana, and
  G.~E. Hinton.
\newblock Neural additive models: {I}nterpretable machine learning with neural
  nets.
\newblock In M.~Ranzato, A.~Beygelzimer, Y.~Dauphin, P.~S. Liang, and
  J.~{Wortman Vaughan}, editors, {\em Proceedings of the 35th Conference on
  Neural Information Processing Systems (NeurIPS 2021)}, volume~34, pages
  4699--4711. Advances in Neural Information Processing Systems, 2021.

\bibitem{alber2019}
M.~Alber, S.~Lapuschkin, P.~Seegerer, M.~H{{\"a}}gele, K.~T. Sch{{\"u}}tt,
  G.~Montavon, W.~Samek, K.-R. M{{\"u}}ller, S.~D{{\"a}}hne, and P.-J.
  Kindermans.
\newblock i{NN}vestigate neural networks!
\newblock {\em Journal of Machine Learning Research}, 20:93, 2019.

\bibitem{Allan2004}
J.~D. Allan.
\newblock Landscapes and riverscapes: {T}he influence of land use on stream
  ecosystems.
\newblock {\em Annual Review of Ecology, Evolution, and Systematics},
  35:257--284, 2004.

\bibitem{apleyZhu}
D.~W. Apley and J.~Zhu.
\newblock Visualizing the effects of predictor variables in black box
  supervised learning models.
\newblock {\em Journal of the Royal Statistical Society, Series B},
  82:1059--1086, 2020.

\bibitem{cai2023}
Y.~Cai.
\newblock Achieve the minimum width of neural networks for universal
  approximation.
\newblock Technical report, arXiv 2209.11395 [cs.LG], 2023.

\bibitem{carlisle2009}
D.~M. Carlisle, J.~Falcone, and M.~R. Meador.
\newblock Predicting the biological condition of streams: {U}se of geospatial
  indicators of natural and anthropogenic characteristics of watersheds.
\newblock {\em Environmental Monitoring and Assessment}, 151:143--160, 2009.

\bibitem{chastaing2012generalized}
G.~Chastaing, F.~Gamboa, and C.~Prieur.
\newblock Generalized {H}oeffding-{S}obol decomposition for dependent variables
  -- application to sensitivity analysis.
\newblock {\em Electronic Journal of Statistics}, 6:2420--2448, 2012.

\bibitem{chen2016xgboost}
T.~Chen and C.~Guestrin.
\newblock {XGB}oost: {A} scalable tree boosting system.
\newblock In B.~Krishnapuram, M.~Shah, A.~J. Smola, C.~Aggarwal, D.~Shen, and
  R.~Rastogi, editors, {\em Proceedings of the 22nd ACM SIGKDD International
  Conference on Knowledge Discovery and Data Mining (KDD '16)}, pages 785--794,
  New York, 2016. Association for Computing Machinery.

\bibitem{ChesapeakeBayProgram2020}
{Chesapeake Bay Program}.
\newblock 2015-2025 stream health management strategy -- v.3, 2020.
\newblock
  https://d18lev1ok5leia.cloudfront.net/chesapeakebay/documents/2015-2025-Stream-Health-Management-Strategy.pdf,
  accessed July 5, 2024.

\bibitem{cybenko1989}
G.~Cybenko.
\newblock Approximation by superpositions of a sigmoidal function.
\newblock {\em Mathematics of Control, Signals, and Systems}, 2:303--314, 1989.

\bibitem{da2021basics}
S.~{Da Veiga}, F.~Gamboa, B.~Iooss, and C.~Prieur.
\newblock {\em Basics and trends in sensitivity analysis: {T}heory and practice
  in {R}}.
\newblock Society for Industrial and Applied Mathematics, Philadelphia, 2021.

\bibitem{gaston2020}
K.~J. Gaston.
\newblock Global patterns in biodiversity.
\newblock {\em Nature}, 405:220--227, 2000.

\bibitem{groemping}
U.~Gr\"omping.
\newblock Model-agnostic effects plots for interpreting machine learning
  models.
\newblock Technical report, Reports in Mathematics, Physics and Chemistry:
  Department II, Beuth University of Applied Sciences, Berlin, 1/2020, 2020.

\bibitem{hill2017predictive}
R.~A. Hill, E.~W. Fox, S.~G. Leibowitz, A.~R. Olsen, D.~J. Thornbrugh, and
  M.~H. Weber.
\newblock Predictive mapping of the biotic condition of conterminous {U.S.}\@
  rivers and streams.
\newblock {\em Ecological Applications}, 27:2397--2415, 2017.

\bibitem{hooker2007generalized}
G.~Hooker.
\newblock Generalized functional {ANOVA} diagnostics for high-dimensional
  functions of dependent variables.
\newblock {\em Journal of Computational and Graphical Statistics}, 16:709--732,
  2007.

\bibitem{hornik1991}
K.~Hornik.
\newblock Approximation capabilities of multilayer feedforward networks.
\newblock {\em Neural Networks}, 4:251--257, 1991.

\bibitem{hornik1989}
K.~Hornik, M.~Stinchcombe, and H.~White.
\newblock Multilayer feedforward networks are universal approximators.
\newblock {\em Neural Networks}, 2:359--366, 1989.

\bibitem{ismailov2022}
V.~Ismailov.
\newblock A three layer neural network can represent any multivariate function.
\newblock {\em Journal of Mathematical Analysis and Applications}, 523:127096,
  2023.

\bibitem{ismailov2023}
A.~Ismayilova and V.~Ismailov.
\newblock On the {K}olmogorov neural networks.
\newblock {\em Neural Networks}, 176:106333, 2024.

\bibitem{kidger}
P.~Kidger and T.~Lyons.
\newblock Universal approximation with deep narrow networks.
\newblock In J.~Abernethy and S.~Agarwal, editors, {\em Proceedings of the
  Thirty Third Conference on Learning Theory}, volume 125, pages 2306--2327.
  Proceedings of Machine Learning Research, 2020.

\bibitem{adam}
D.~P. Kingma and J.~Ba.
\newblock {A}dam: {A} method for stochastic optimization.
\newblock Technical report, arXiv 1412.6980v9 [cs.LG], 2017.

\bibitem{lengerich}
B.~Lengerich, S.~Tan, C.-H. Chang, G.~Hooker, and R.~Caruana.
\newblock Purifying interaction effects with the functional {ANOVA}: {A}n
  efficient algorithm for recovering identifiable additive models.
\newblock In S.~Chiappa and R.~Calandra, editors, {\em Proceedings of the
  Twenty Third International Conference on Artificial Intelligence and
  Statistics}, volume 108, pages 2402--2412. Proceedings of Machine Learning
  Research, 2020.

\bibitem{lepore:2022}
A.~Lepore, B.~Palumbo, and J.-M. Poggi, editors.
\newblock {\em Interpretability for Industry 4.0: {S}tatistical and Machine
  Learning Approaches}.
\newblock Springer, Cham, 2022.

\bibitem{linardatos2020explainable}
P.~Linardatos, V.~Papastefanopoulos, and S.~Kotsiantis.
\newblock Explainable {AI}: {A} review of machine learning interpretability
  methods.
\newblock {\em Entropy}, 23:18, 2021.

\bibitem{luenberger1969}
D.~G. Luenberger.
\newblock {\em Optimization by Vector Space Methods}.
\newblock Wiley, New York, 1969.

\bibitem{maloney2022explainable}
K.~O. Maloney, C.~Buchanan, R.~D. Jepsen, K.~P. Krause, M.~J. Cashman, B.~P.
  Gressler, J.~A. Young, and M.~Schmid.
\newblock Explainable machine learning improves interpretability in the
  predictive modeling of biological stream conditions in the {C}hesapeake {B}ay
  {W}atershed, {USA}.
\newblock {\em Journal of Environmental Management}, 322:116068, 2022.

\bibitem{miller2019explanation}
T.~Miller.
\newblock Explanation in artificial intelligence: {I}nsights from the social
  sciences.
\newblock {\em Artificial Intelligence}, 267:1--38, 2019.

\bibitem{molnar2020interpretable}
C.~Molnar.
\newblock {\em Interpretable Machine Learning -- A Guide for Making Black Box
  Models Explainable}.
\newblock Independently published, 2 edition, 2022.

\bibitem{murdoch2019}
W.~J. Murdoch, C.~Singh, K.~Kumbier, R.~Abbasi-Asl, and B.~Yu.
\newblock Definitions, methods, and applications in interpretable machine
  learning.
\newblock {\em Proceedings of the National Academy of Sciences of the United
  States of America}, 116:22071--22080, 2019.

\bibitem{namayandeh2018}
A.~Namayandeh, D.~V. Beresford, K.~M. Somers, and P.~J. Dillon.
\newblock Difference in benthic invertebrate communities of headwater streams
  can be detected using a short elevation gradient.
\newblock {\em International Aquatic Research}, 10:153--164, 2018.

\bibitem{park2021minimum}
S.~Park, C.~Yun, J.~Lee, and J.~Shin.
\newblock Minimum width for universal approximation.
\newblock In {\em Proceedings of the Ninth International Conference on Learning
  Representations}. OpenReview.net, 2021.
\newblock https://openreview.net/forum?id=O-XJwyoIF-k.

\bibitem{ruegamer2023}
D.~R\"ugamer.
\newblock A new {PHO}-rmula for improved performance of semi-structured
  networks.
\newblock In A.~Krause, E.~Brunskill, K.~Cho, B.~Engelhardt, S.~Sabato, and
  J.~Scarlett, editors, {\em Proceedings of the 40th International Conference
  on Machine Learning}, volume 202, pages 29291--29305. Proceedings of Machine
  Learning Research, 2023.

\bibitem{sarker2021machine}
I.~H. Sarker.
\newblock Machine learning: {A}lgorithms, real-world applications and research
  directions.
\newblock {\em SN Computer Science}, 2:160, 2021.

\bibitem{Schmidhuber2015DNN}
J.~Schmidhuber.
\newblock Deep learning in neural networks: {A}n overview.
\newblock {\em Neural Networks}, 61:85--117, 2015.

\bibitem{shen2022}
Z.~Shen, H.~Yang, and S.~Zhang.
\newblock Optimal approximation rate of {ReLU} networks in terms of width and
  depth.
\newblock {\em Journal de Math\'ematiques Pures et Appliqu\'ees}, 157:101--135,
  2022.

\bibitem{smith2017refinement}
Z.~M. Smith, C.~Buchanan, and A.~Nagel.
\newblock Refinement of the {B}asin-wide {I}ndex of {B}iotic {I}ntegrity for
  non-tidal streams and wadeable rivers in the {C}hesapeake {B}ay {W}atershed.
\newblock {\em Interstate Commission on the Potomac River Basin Report}, 17-2,
  2017.

\bibitem{snyder2003}
C.~D. Snyder, J.~A. Young, R.~Villella, and D.~P. Lemari\'e.
\newblock Influences of upland and riparian land use patterns on stream biotic
  integrity.
\newblock {\em Landscape Ecology}, 18:647--664, 2003.

\bibitem{stone1994}
C.~J. Stone.
\newblock The use of polynomial splines and their tensor products in
  multivariate function estimation.
\newblock {\em The Annals of Statistics}, 22:118--171, 1994.

\bibitem{tibshiraniLasso}
R.~Tibshirani.
\newblock Regression shrinkage and selection via the lasso.
\newblock {\em Journal of the Royal Statistical Society, Series B},
  58:267--288, 1996.

\bibitem{welchowski}
T.~Welchowski, K.~O. Maloney, R.~Mitchell, and M.~Schmid.
\newblock Techniques to improve ecological interpretability of black-box
  machine learning models -- case study on biological health of streams in the
  {U}nited {S}tates with gradient boosted trees.
\newblock {\em Journal of Agricultural, Biological and Environmental
  Statistics}, 27:175--197, 2022.

\bibitem{Zeiler2014Visualizing}
M.~D. Zeiler and R.~Fergus.
\newblock Visualizing and understanding convolutional networks.
\newblock In D.~Fleet, T.~Pajdla, B.~Schiele, and T.~Tuytelaars, editors, {\em
  Proceedings of the 13th European Conference on Computer Vision (ECCV 2014)},
  volume 8689 of {\em Lecture Notes in Computer Science}, pages 818--833, Cham,
  2014. Springer International Publishing.

\end{thebibliography}

\section*{Appendix}

\subsection*{A \ Details on NAM fitting} 
As stated above, each function $f_\theta^0$ in~\reff{eq:NAM} is represented by a separate ANN. This representation is generally not restricted to a specific network architecture but can be adapted to the learning task(s) as needed. For our numerical experiments, we used fully connected ANNs with five hidden layers each. The numbers of units were $256$, $128$, $64$, $32$, and $8$, starting with the first hidden layer. Rectified linear unit activation functions were used in the first four hidden layers; a linear activation function was used in the last hidden layer with $b_\theta = 8$. Twenty percent dropout was applied to the second, third and fourth hidden layers. The NAM was fitted using backpropagation with the mean squared error loss and the Adam optimizer~\cite{adam}. The backpropagation procedure was run until convergence. No regularization techniques were used except dropout. 

\subsection*{B \ Details on post-hoc orthogonalization} In Step 3 of our method, we apply the following algorithm to process the initial intercept estimate $\mu^0 = f_\emptyset^0=0$ and the NAM estimates $f_\theta^0$, $\theta \in \mathcal{P}(\Upsilon)\backslash\emptyset$:\\

\noindent {Input:} Vectors of initial estimates $\mathbf{f}_\theta^0 = \mathbf{U}_\theta \mathbf{w}_\theta^0 \in \mathbb{R}^n$, $\theta \in \mathcal{P}(\Upsilon)$.\\ 

\noindent For $m=1$ to $d-1$:
\begin{enumerate}
\item[1.1] Define the {\em actual} interaction order by $d-m+1$.
\item[1.2] Define the {\em actual} set of effects by $\mathcal{A}=\{\theta \in \mathcal{P}(\Upsilon): |\theta| = d-m+1\}$. Let ${{f}}_{\mathcal{A}}^{m-1} = \{ \mathbf{{f}}_\theta^{m-1} \}_{\theta \in \mathcal{A}}$ be the set of function estimates of order $d-m+1$.
\item[1.3] Define the set of {\em lower-order} effects by $\mathcal{L}=\{\theta \in \mathcal{P}(\Upsilon): |\theta| < d-m+1\}$. Let ${{f}}_{\mathcal{L}}^{m-1} = \{ \mathbf{{f}}_\theta^{m-1} \}_{\theta \in \mathcal{L}}$ be the set of function estimates of order lower than $d-m+1$.
\item[1.4] Define the set of {\em higher-order} effects by $\mathcal{H}=\{\theta \in \mathcal{P}(\Upsilon): |\theta| > d-m+1\}$. Let ${{f}}_{\mathcal{H}}^{m-1} = \{ \mathbf{{f}}_\theta^{m-1} \}_{\theta \in \mathcal{H}}$ be the set of function estimates of order higher than $d-m+1$.
\item[1.5] Define the matrix $\mathbf{U} = [ \mathbf{U}_\theta ]_{\theta \in \mathcal{L}}$ by concatenating the output matrices $\mathbf{U}_\theta$, $\theta\in \mathcal{L}$ (including the single-column matrix $\mathbf{U}_\emptyset=(1,\ldots , 1)^\top$ for the intercept).
By definition, $\mathbf{U}$ is of dimension $n \times B$, where $B = \sum_{\theta \in \mathcal{L}} b_\theta$. We assume that the architectures of the ANN terms in \reff{eq:NAM} have been specified such that $n \ge B$.
\end{enumerate}
\begin{enumerate}
\item[2.1] Compute the matrix $\mathbf{P} = \mathbf{U} (\mathbf{U}^\top \mathbf{U})^{-1} \mathbf{U}^\top$ (assuming $\mathbf{U}$ is of full rank). By definition, multiplication of a vector $\mathbf{x} \in \mathbb{R}^n$ with $\mathbf{P}$ is equivalent to projecting $\mathbf{x}$ onto the column space spanned by $\mathbf{U}$. In case $\mathbf{U}$ is not of full rank, we adapt the algorithm as described below.
\item[2.2] Compute the sum of the actual function estimates by $\mathbf{{z}}_{\mathcal{A}}^{m-1} = \sum_{\theta \in \mathcal{A}} \mathbf{{f}}_\theta^{m-1}$.
\end{enumerate}
\begin{enumerate}
\item[3.1] Update the actual effects ${{f}}_{\mathcal{A}}^m$ by projecting $\mathbf{{z}}_{\mathcal{A}}^{m-1}$ onto the column space of~$\mathbf{U}$ and by setting ${{f}}_{\mathcal{A}}^m$ equal to the vectors that are orthogonal to this projection. This gives ${{f}}_{\mathcal{A}}^m = \{ (\mathbf{I}-\mathbf{P}) \mathbf{{f}}_\theta^{m-1} \}_{\theta \in \mathcal{A}}$, where $\mathbf{I}$~is the identity matrix of size $n$.
\item[3.2] Update the lower-order effects ${{f}}_{\mathcal{L}}^m$ by adding the projections of $\mathbf{{z}}_{\mathcal{A}}^{m-1}$ to~${{f}}_{\mathcal{L}}^{m-1}$. This gives
${{f}}_{\mathcal{L}}^m = \{ \mathbf{{f}}_\theta^{m-1} + \mathbf{U}_\theta [(\mathbf{U}^\top \mathbf{U})^{-1} \mathbf{U}^\top \mathbf{{z}}_{\mathcal{A}}^{m-1}]_{ \theta} \}_{\theta \in \mathcal{L}} $, where the vector  $[(\mathbf{U}^\top \mathbf{U})^{-1} \mathbf{U}^\top \mathbf{{z}}_{\mathcal{A}}^{m-1}]_{ \theta}$ contains those elements of $(\mathbf{U}^\top \mathbf{U})^{-1} \mathbf{U}^\top \mathbf{{z}}_{\mathcal{A}}^{m-1}$ matching the positions of the columns of~$\mathbf{U}_\theta$ in~$\mathbf{U}$. 
\item[3.3] The higher-order effects are not updated, i.e.\@ ${{f}}_{\mathcal{H}}^{m} = {{f}}_{\mathcal{H}}^{m-1} = \{ \mathbf{{f}}_\theta^{m-1} \}_{\theta \in \mathcal{H}}$.
\end{enumerate}
The updates in Step 3.2 imply that each
$\mathbf{f}_\theta^{m}$ can be written in the form $\mathbf{U}_\theta \mathbf{\beta}_\theta^m$, where $\mathbf{\beta}_\theta^m$ is a vector of coefficients of length~$b_\theta$. Consequently, one obtains
\begin{eqnarray} \label{orthoProof}
&&\hspace*{-.6cm}\Big(\sum_{\theta \in \mathcal{L}} \mathbf{f}_\theta^m
\Big)^\top 
\sum_{\theta \in \mathcal{A}} \mathbf{f}_\theta^m \nonumber\\
&&\hspace*{-.6cm} =
\Big(
\sum_{\theta \in \mathcal{L}} \mathbf{U}_\theta \mathbf{\beta}_\theta^{m -1}  +
\mathbf{U} (\mathbf{U}^\top \mathbf{U})^{-1} \mathbf{U}^\top \mathbf{{z}}_{\mathcal{A}}^{m-1}
\Big)^\top
(\mathbf{I}-\mathbf{P})\, \mathbf{{z}}_{\mathcal{A}}^{m-1} \nonumber\\
&& \hspace*{-.6cm} =
\big( [\mathbf{\beta}_\theta^{m-1}]_{\theta\in\mathcal{L}} \big)^\top \mathbf{U}^\top
(\mathbf{I}-\mathbf{P})\, \mathbf{{z}}_{\mathcal{A}}^{m-1} + \mathbf{{z}}_{\mathcal{A}}^{m-1,\top} \, \mathbf{P} \, (\mathbf{I}-\mathbf{P})\, \mathbf{{z}}_{\mathcal{A}}^{m-1} \nonumber\\[.1cm]
&& \hspace*{-.6cm} = 0 \, ,
\end{eqnarray}
where $[\mathbf{\beta}_\theta^{m-1}]_{\theta\in\mathcal{L}}$ denotes the concatenation of the coefficient vectors $\mathbf{\beta}_\theta^{m-1}$ (i.e., a vector of length $B$). According to \reff{orthoProof}, the sum of the lower-order effects is orthogonal to the sum of actual effects, and the final result of the algorithm satisfies the stacked orthogonality constraints in \reff{eq:stackedOrthog}.

In the final step, we center the vectors $\mathbf{f}_\theta^{d-1}$, $\theta\in \mathcal{P}(\Upsilon)\backslash\emptyset$, by subtracting their respective means. This ensures that all functions are centered around zero, as assumed in the Subsection ``Conditions on the features and the prediction function''. Note that the centering does not affect the above orthogonality proof, as the actual effects $\mathbf{f}_\theta^m$, $\theta \in\mathcal{A}$, are left unchanged in later iterations (implying $\mathbf{f}_\theta^m = \mathbf{f}_\theta^{d-1}$ for these effects), and as the sum of the {\em mean-centered} actual effects is equal to the sum of the {\em uncentered} actual effects $\sum_{\theta \in \mathcal{A}} \mathbf{f}_\theta^m$ in the first line of \reff{orthoProof}. The latter result is due to the fact that the sum $\sum_{\theta \in \mathcal{A}} \mathbf{f}_\theta^m$ has zero mean, being orthogonal to $\mathbf{U}_\emptyset = (1,\ldots , 1)^\top$. By the same argument, the centering does not affect the value of the intercept term.

In case $\mathbf{U}$ is not of full rank, 
we project $\mathbf{z}_\mathcal{A}^{m-1}$ onto a full-rank subspace of the column space of $\mathbf{U}$.
More specifically, we consider the pivoted QR decomposition
\begin{equation}
\mathbf{U} = \tilde{\mathbf{Q}} \, \tilde{\mathbf{R}} \, \tilde{\mathbf{P}}^\top \, ,
\end{equation}
where $\tilde{\mathbf{Q}} \in \mathbb{R}^{n\times n}$ is a unitary matrix, $\tilde{\mathbf{R}} \in \mathbb{R}^{n\times B}$ is an upper triangular matrix with diagonal elements $r_{11},\ldots , r_{B B}$, and $\tilde{\mathbf{P}} \in \mathbb{R}^{B\times B}$ is a permutation matrix arranging the columns of~$\mathbf{U}$ such that $|r_{11}| \ge \ldots \ge |r_{B B}|$. Denoting the rank (i.e., the number of non-zero singular values) of $\mathbf{U}$ by $r_\mathbf{U}$, we define $\tilde{\mathbf{U}} \in \mathbb{R}^{n \times r_\mathbf{U}}$ by those columns of $\mathbf{U}$ corresponding to first $r_\mathbf{U}$ diagonal elements of~$\tilde{\mathbf{R}}$. The positions of these columns are indicated by the entries of the permutation matrix $\tilde{\mathbf{P}}$. Accordingly, we define the matrices $\tilde{\mathbf{U}}_\theta$, $\theta \in\mathcal{L}$, by those columns of $\mathbf{U}_\theta$ contained in $\tilde{\mathbf{U}}$, and we perform Steps~2 and~3 of the above algorithm with $\mathbf{U}$ and $\mathbf{U}_\theta$ replaced by $\tilde{\mathbf{U}}_\theta$ and~$\tilde{\mathbf{U}}_\theta$, respectively.

\subsection*{C \ Experiments with synthetic data -- definition of the main effects and the two-way interactions}

As stated in the main text, we considered three sets of functional forms for the effects $f_1,f_2,f_3,f_{12},f_{13},f_{23}$. These were defined as follows:\\

\noindent {\bf Scenario 1}
\begin{itemize}
\item ${f_1(X_1)} = \cos (2 \cdot X_1)$
\item ${f_2(X_2}) = \tanh (0.5 \cdot X_2)$
\item ${f_3(X_3}) = d_{\text{norm}}(X_3-1.5)+d_{\text{norm}}(X_3+1.5)$
\item ${f_{12}(X_1,X_2)} = \sin(1.5\cdot (X_1^2-X_2^2)+d_{\text{norm}}(0.5 \cdot X_1 \cdot X_2)$
\item ${f_{13}(X_1,X_3)} = \cos(X_{1} + X_3) + \sin(X_{1} \cdot X_3)$
\item ${f_{23}(X_2,X_3)} = 0.5\cdot\sin((X_2-1)^2+(X_3+1)^2)$
\end{itemize}

\vspace{.1cm}
\noindent {\bf Scenario 2}
\begin{itemize}
\item ${f_1(X_1)} = -3\cdot\cos(3 \cdot X_1-2)\cdot X_{1} $
\item ${f_2(X_2}) = p_{\text{Weibull}, 3, 1}(X_2) \cdot \mathrm{I} (X_2 \geq 0) +
        p_{\text{Weibull},0.5, 1}(-X_2) \cdot \mathrm{I} (X_2 < 0)$
\item ${f_3(X_3}) = X_3^2$
\item ${f_{12}(X_1,X_2)} = \sin(0.75\cdot X_1\cdot X_2) + 0.25\cdot\sqrt{X_1^2+X_2^2}+0.25\cdot\cos((X_1-\pi)\cdot(X_2+\pi))$
\item ${f_{13}(X_1,X_3)} = \sin(X_1^2+X_3^2)$
\item ${f_{23}(X_2,X_3)} = \sin(X_2+X_3)$
\end{itemize}

\vspace{.1cm}
\noindent {\bf Scenario 3}
\begin{itemize}
\item ${f_1(X_1)} = \cos(2\cdot X_1)$
\item ${f_2(X_2}) = \tanh(X_2)$
\item ${f_3(X_3}) = -X_3^3$
\item ${f_{12}(X_1,X_2)} = -\cos(1.5\cdot X_1 - 0.75\cdot X_2)$
\item ${f_{13}(X_1,X_3)} = \cos((X_1-\pi)\cdot(X_3+\pi))$
\item ${f_{23}(X_2,X_3)} = 4\cdot (d_{\text{norm}}(X_{2})/d_{\text{norm}}(0) - 0.5)\cdot(p_{\text{norm}}(X_3)-0.5)+\sin(X_2+X_3)$
\end{itemize} \vspace{.35cm}
\noindent The ten-way interaction $f_{1,\ldots,10} \, (X_{1},\ldots ,X_{10})$ was the same in all scenarios. It was defined by $f_{1,\ldots,10}(X_{1},\ldots ,X_{10})=2\cdot (
(\prod_{j=1}^{10}|X_{j}| / 1000)^{1/8}-0.5)$.

\vspace{.4cm}

\noindent Notes:
\begin{itemize}
\item[(i)] $p_\text{Weibull,$\alpha$,$\sigma$}$ refers to the c.d.f.\@ of the Weibull distribution with shape parameter~$\alpha$ and scale parameter $\sigma$. It is defined by
$p_\text{Weibull,$\alpha$,$\sigma$}(x)=(\alpha/\sigma) \cdot (x/\sigma)^{\alpha-1} \cdot \exp {(-(x/\sigma)^{\alpha})}$.\vspace{-.15cm}
\item[(ii)] $p_\text{norm}$ and $d_{\text{norm}}$ refer to the c.d.f.\@ and the p.d.f.\@ of the standard normal distribution, respectively.
\end{itemize}

\newpage

\subsection*{D \ Experiments with synthetic data -- additional results}

\begin{figure}[h]
    \centering
    \includegraphics[width = 0.6\textwidth]{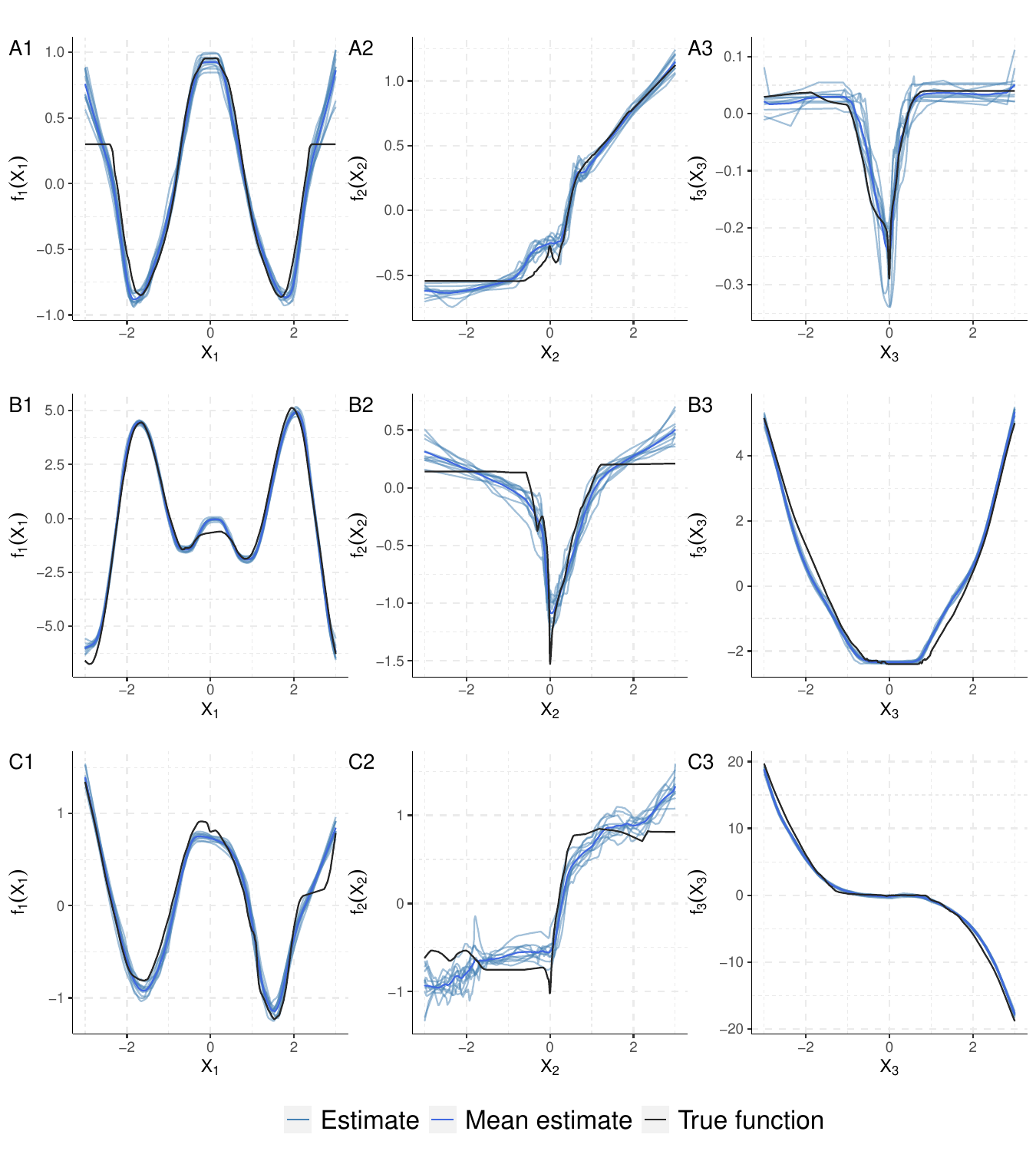}
    \caption{Experiments with synthetic data. The blue lines visualize the main effects $f_1(X_1), f_2(X_2), f_3(X_3)$, as obtained by applying the proposed three-step algorithm to samples of size $n=5000$. The black lines correspond to the true post-hoc-orthogonalized main effects defined in Appendix C (A1-A3: scenario 1, B1-B3: scenario 2, C1-C3: scenario 3).}
     \label{sim:fig5000main}
\end{figure}

\begin{figure}
    \centering
    \includegraphics[width = 1\textwidth]{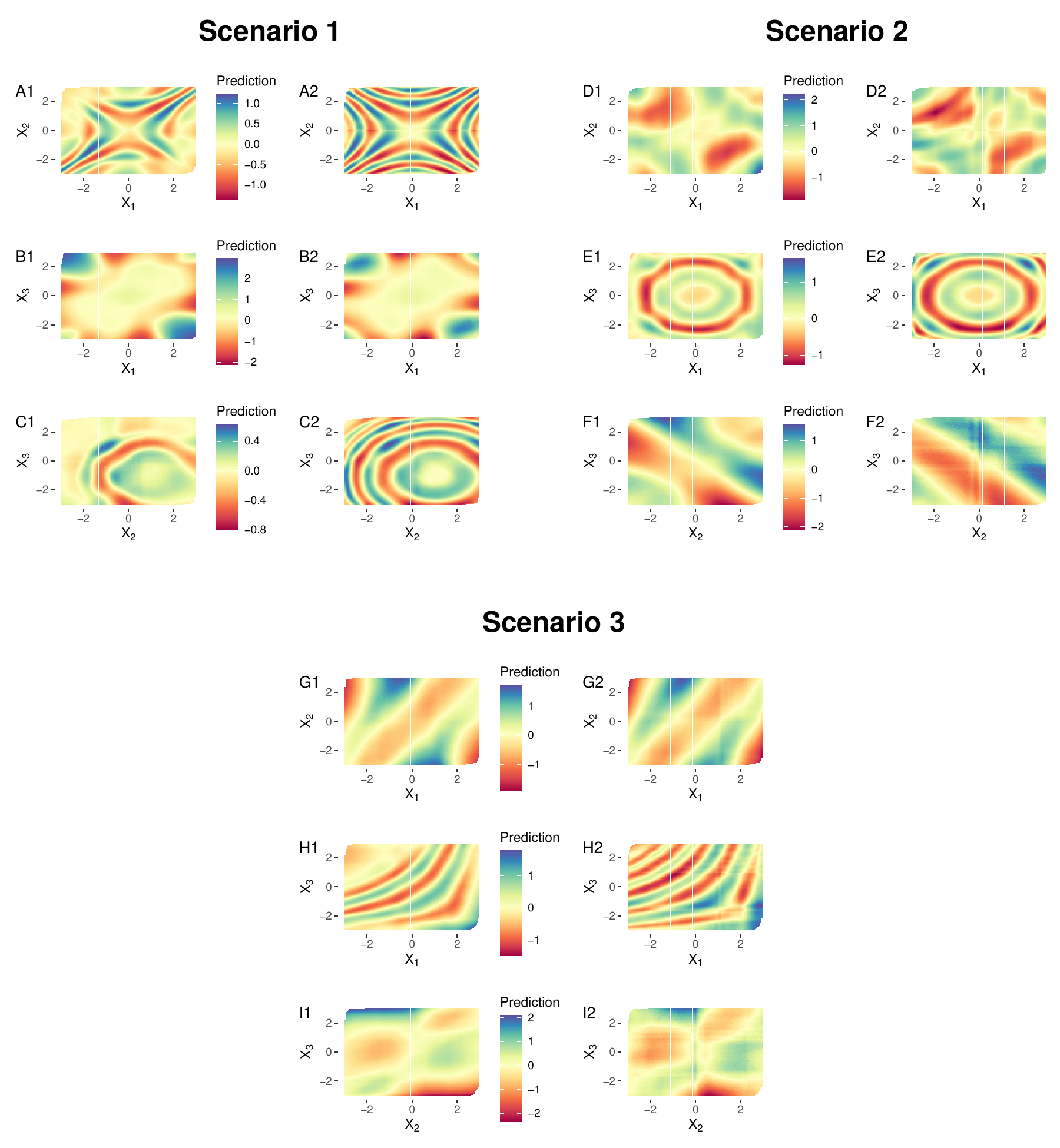}
    \caption{Experiments with synthetic data. The left column presents the average two-way interaction effects $f_{12}(X_{12}), f_{13}(X_{13}), f_{23}(X_{23})$, as obtained by applying the proposed three-step algorithm to samples of size $n=2000$. The right columns contain the true post-hoc-orthogonalized two-way interactions defined in Appendix~C.}
     \label{sim:fig2000inter}
\end{figure}

\begin{figure}
    \centering
    \includegraphics[width = 1\textwidth]{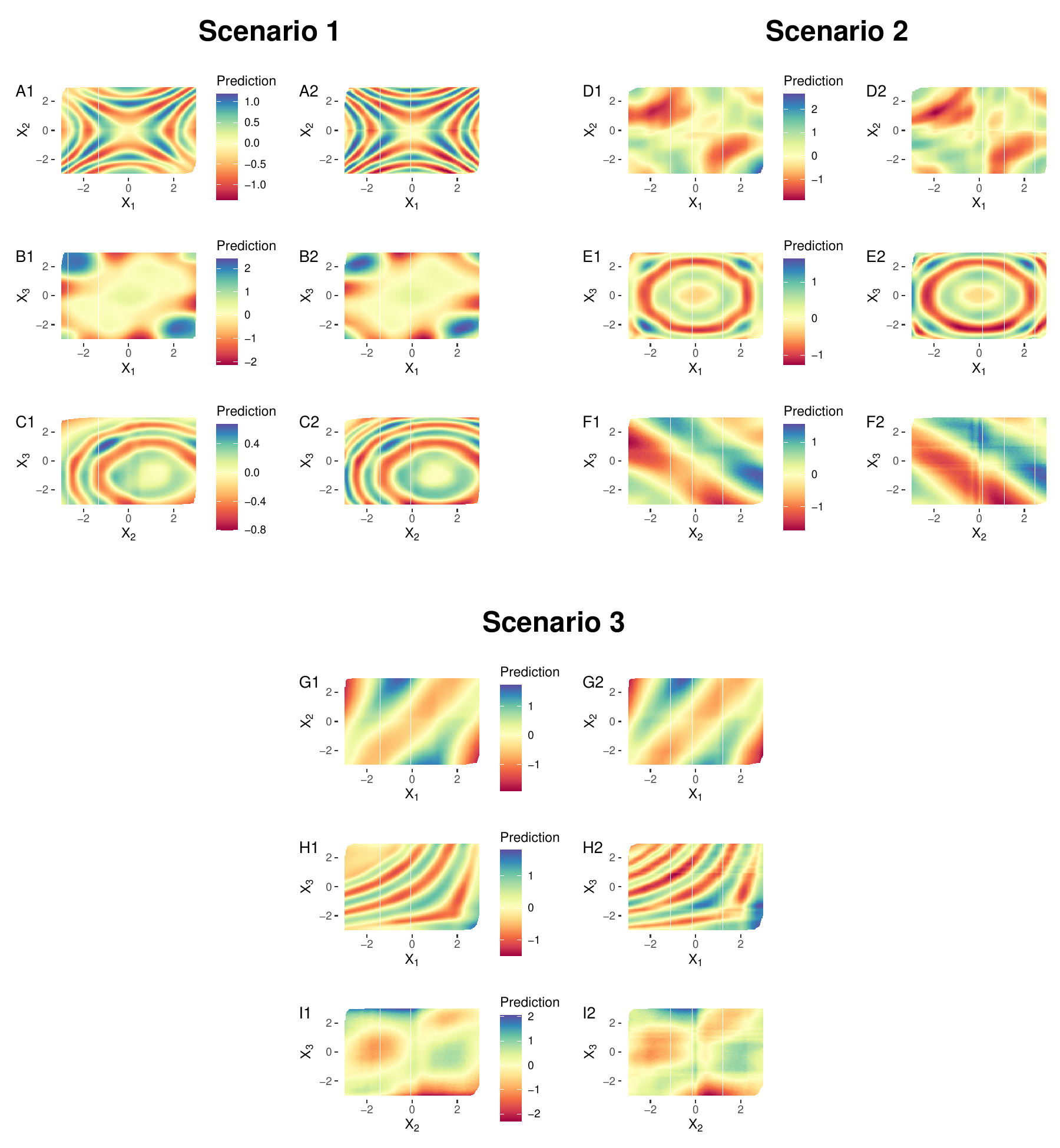}
    \caption{Experiments with synthetic data. The left column presents the average two-way interaction effects $f_{12}(X_{12}), f_{13}(X_{13}), f_{23}(X_{23})$, as obtained by applying the proposed three-step algorithm to samples of size $n=5000$. The right columns contain the true post-hoc-orthogonalized two-way interactions defined in  Appendix~C.}
     \label{sim:fig5000inter}
\end{figure}

\end{document}